\documentclass{svproc}

\usepackage{graphicx}
\usepackage{amsmath,amssymb,amsfonts}
\usepackage[labelfont=bf]{caption}
\usepackage{subcaption}
\usepackage{threeparttable}
\usepackage{multirow}
\usepackage[leftcaption]{sidecap} 
\sidecaptionvpos{t}{figure}
\usepackage{url}

\usepackage{enumitem}
\setlist[itemize,1]{label=\textbullet}

\begin{document}

\title{An Application of Deep Learning for Sweet Cherry Phenotyping using YOLO Object Detection}

\titlerunning{Deep Learning for Sweet Cherry Phenotyping using YOLO Object Detection}  

\author{Ritayu Nagpal, Sam Long, Shahid Jahagirdar, Weiwei Liu, \\ Scott Fazackerley, Ramon Lawrence\inst{1} \and Amritpal Singh\inst{2}}

\authorrunning{Singh et al.} 
%
\tocauthor{Ritayu Nagpal, Sam Long, Shahid Jahagirdar, Weiwei Liu, Scott Fazackerley, Ramon Lawrence, Amritpal Singh}

\index{Nagpal, R.}
\index{Long, S.}
\index{Jahagirdar, S.}
\index{Liu, W.}
\index{Fazackerley, S.}
\index{Lawrence, R.}
\index{Singh, A.}

\institute{
Department of Computer Science, University of British Columbia \\ Kelowna, BC, Canada, V1V 2Z3\\
\email{ritayu.nagpal09@gmail.com, lsam8910@gmail.com, \\ shahid.h.j@gmail.com, weiwei.liu2046@gmail.com, scott.fazackerley@alumni.ubc.ca, ramon.lawrence@ubc.ca}
\and
Agriculture and Agri-Food Canada\\
Summerland, BC, Canada, V0H 1Z0\\
\email{amritpal.singh@agr.gc.ca}
}

\maketitle

\begin{abstract}
Tree fruit breeding is a long-term activity involving repeated measurements of various fruit quality traits on a large number of samples. These traits are traditionally measured by manually counting the fruits, weighing to indirectly measure the fruit size, and fruit colour is classified subjectively into different color categories using visual comparison to colour charts. These processes are slow, expensive and subject to evaluators’ bias and fatigue. Recent advancements in deep learning can help automate this process. Objective data can be generated for consistent characterization of germplasm, with greater speed and higher accuracy. 
A method was developed to automatically count the number of sweet cherry fruits in a camera’s field of view in real time using YOLOv3. A system capable of analyzing the image data for other traits such as size and color was also developed using Python. The YOLO model obtained close to 99\% accuracy in object detection and counting of cherries and 90\% on the Intersection over Union metric for object localization when extracting size and colour information.  The model surpasses human performance and offers a significant improvement compared to manual counting.
\end{abstract}


\section{Introduction}

The Okanagan Valley in British Columbia, Canada is one of the largest cherry producing regions in the country~\cite{Neilen:2014} accounting for over 89\% of Canada's production of sweet cherries~\cite{nelson_2018}. The Agriculture and Agri-Food Canada (AAFC) research station located in Summerland, BC, has produced many of the premier cherry varieties currently in commercial production and has a large, ongoing research program to develop new varieties with traits such as improved fruit quality, later harvest, and self-fertility~\cite{Agrifoods:2013}. New improved cultivars of fruits produced through selective breeding for desirable traits are known to significantly increase the economic growth and success of the horticulture industry. 

The process of tree fruit breeding involves development of crosses (hybridizations) using distinct genotypes, followed by several stages of evaluation of the germplasm for numerous traits over multiple years and locations before the commercialization of a new cultivar. Germplasm is a piece of live plant tissue from which a new plant can be grown and is a common method used for crop propagation.  Due to the long juvenile period and perennial nature of trees, this process can take more than 20 years. The resources and time required for evaluation of the germplasm for various traits is a major bottleneck in this process.

Traditionally, in each phase of germplasm evaluation, most of the traits are measured manually, which involves tedious and repetitive tasks such as counting fruits, size measurement, and color classification.  The assessment of these traits is referred to as phenotyping. The tree fruit germplasm evaluation process is therefore vulnerable to worker fatigue and subjectivity, resulting in errors and inconsistencies in the collection of fruit trait data.

With recent advancements in image analysis, computer vision and artificial intelligence, many of the aforementioned manual tasks involved in the process of tree fruit breeding can be fully or partially automated.  
Open source tools and methods were explored and a Python-based application utilizing YOLO (You Only Look Once) and OpenCV for real-time sweet cherry detection was developed allowing for increased productivity and consistency in analysis. 

A variety of open-source object detection models were considered. These models varied from traditional image processing techniques such as Haar-Feature based classifiers to more modern deep learning architectures such as Fast-R-CNN, YOLO (You Only Look Once) and SSDs (Single-Shot Detectors). Techniques such as transfer learning were employed to train the models quickly and accurately on the labelled custom dataset. The final model was integrated with Tkinter and OpenCV libraries in Python to make a full-fledged desktop application with all the image statistics and analysis results such as size, color, and counts collected and stored in a spreadsheet for further analysis. 

Automation can help breeding programs in improving the overall efficiency of the evaluation process by providing unbiased and precise data in a swift manner while avoiding costs arising from tedious procedures and possible human errors. Under controlled environments, the application is capable of generating accurate counts of sweet cherries, as well as their respective size and color information in an instant, surpassing the performance of humans on the same task in terms of both time and accuracy. 

The contributions of this work are:

\begin{itemize}
    \item Evaluation of numerous image processing classifiers on the cherry detection and classification problem
    
    \item Development of an integrated application for statistics collection using the image processing classifier
    
    \item Analysis of the benefits of automation and the real-world impact on development of cherry cultivars in breeding programs
    
    \item A complete application that offers significant improvements in time, data quality and testing results for cherry phenotyping. 
    
\end{itemize}

The paper outline is as follows. Section~\ref{sec:bg} discusses current techniques used for cherry phenotyping and presents different deep learning approaches for object classification. The dataset, model training and model selection are in Section~\ref{sec:modelDev}. Information on models and analysis is in Section~\ref{sec:Analysis}. The paper closes with future work and conclusions.

\section{Background}\label{sec:bg}

Properties such as fruit size and color of each genotype of cherry are among the key factors in the evaluation and selection of cherry germplasm. Traditionally, these properties are evaluated manually. The following steps are utilized:

\begin{itemize}
    \item Two sets of 100 cherries are counted from a selected genotype, and weighed to produce an estimation of fruit size in terms of average fruit weight.
    \item Cherries are assigned a color number on a scale of 1 to 7 by an evaluator based on a CTIFL standard cherry color card\footnote{https://www.ctifl.fr}.
    \item A sub-sample of cherries are put onto a firmness tester instrument in batches to measure the firmness and size distribution of each cherry.
\end{itemize}

Ensuring each cherry sample comprises exactly 100 cherries is crucial. The same amount of cherries ensures each genotype is evaluated on the same base line in the estimation of size and weight. Traditionally, cherries are counted manually one by one. This process is time consuming and prone to miscounts. It takes about 60 sec to count a sample of 100 cherries.

\begin{figure*}[!t]
\captionsetup[subfigure]{justification=centering}
\begin{subfigure}[t]{0.5\textwidth}
\centering
\includegraphics[width=.9\textwidth]{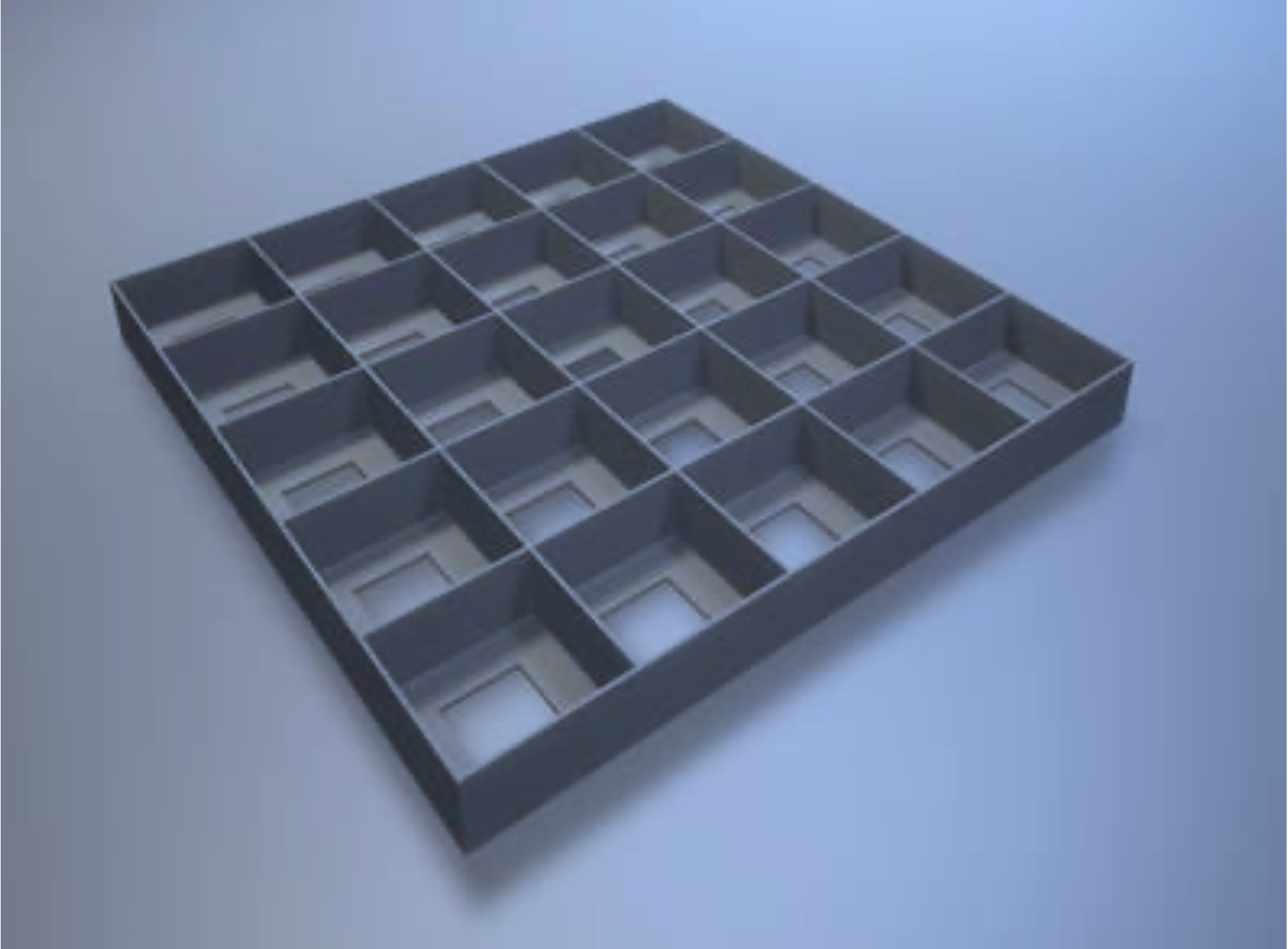}
\caption{Cherry count tray }
\label{fig:cherryCountTray}
\end{subfigure}
\hfill
\begin{subfigure}[t]{0.5\textwidth}
\centering
\includegraphics[width=.6\textwidth]{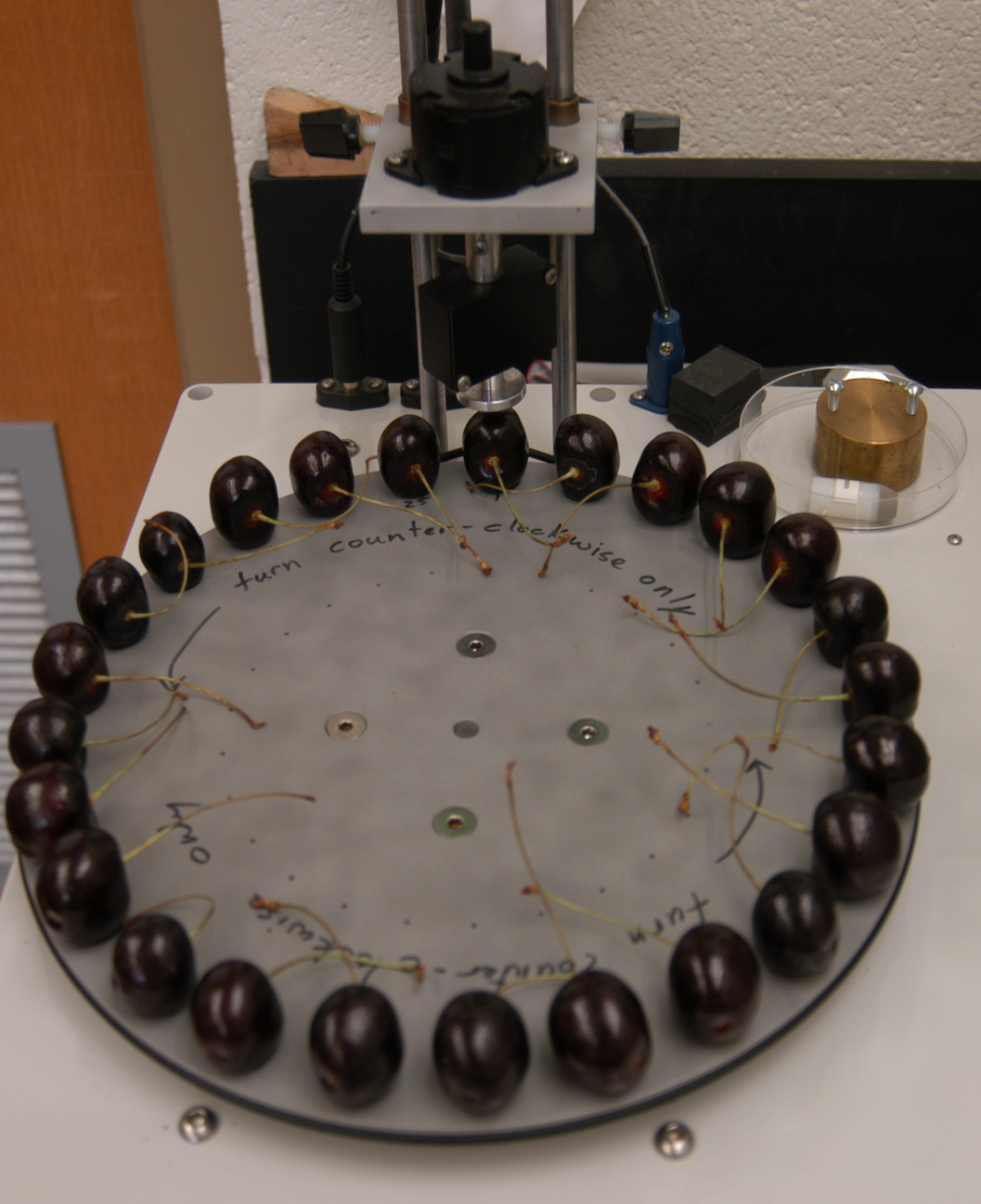}
\caption{Cherry size and firmness measurement machine}
\label{fig:cherrySize}
\hfill
\end{subfigure}
\caption{Cherry measurement apparatus}
\end{figure*}

To make this process more efficient, AAFC introduced a 3D printed tray with 25 cubicles (Figure~\ref{fig:cherryCountTray}). Without counting, the operator only needs to put one cherry in each cubicle, and as long as all cubicles are filled, the tray will contain 25 cherries. Repeating the process four times results in a 100 cherry sample. This method has eliminated the requirement of counting cherries one by one, but it still takes more than 20 seconds to finish loading 4 batches of 25 cherries in a weighing tray. This method improved the accuracy, however, there can still be some accidental errors in getting exactly 100 cherries.

To evaluate the color of each cherry sample, the evaluator observes the cherries collectively, and then compares the color of the cherries with the CTIFL cherry color chart to classify the fruits. There are seven classes of cherry colors in the CTIFL colour chart, and each cherry sample is categorized into one or two classes. The major drawback of this evaluation process is that the evaluation is subjective, as the samples could be categorized into different color categories by different evaluators, or by the same evaluator at a different time.  This method is further impacted by ambient lighting conditions and evaluator fatigue.

One method to estimate the cherry size is through weight averaging. Since the density of each cherry is approximately the same, the overall size of each cherry genotype is estimated by weight index. To get an accurate estimate of cherry size, the diameter of each fruit is measured using an instrument that also measures the fruit firmness (Figure~\ref{fig:cherrySize}). 

Object recognition is used to describe a collection of related computer vision tasks that involve identifying objects in digital images/videos. An object detection model identifies known objects (objects a model is trained on) and outputs information about their positions in terms of bounding box coordinates. 

Image classification is a task involving the prediction of the object class for input by providing a probabilistic output for all the classes available. Object localization refers to determining the location of one or more objects in an image and drawing a bounding box around these identified objects. Another method for object localization is by marking/classifying all the pixels of an image and generating an image mask. The later method for object localization is also referred to as object segmentation. Object detection performs both these tasks and localizes and classifies one or more objects in an image.

An object detection model is trained by providing images and labels corresponding to those images having classification and location information.  Models can be trained with images that contain various types of fruits, along with a label that specifies the location of objects in the image and class of fruit they represent. Given an image, the model outputs a list of the objects detected, the location of a bounding box that contains each object, and a score that indicates the confidence that detection was correct.

Several object detection models have been developed over the years, and this work evaluates the most suitable object detection models for the cherry detection problem. Most of the modern modeling architectures fall under One-Step or Two-Step Object Detection techniques~\cite{objectDetection:2020}. 

Two-Step Object Detection involves algorithms that first identify bounding boxes that may potentially contain objects and then classify each bounding box separately in the second step. The first step requires a Region Proposal Network, providing many regions that are then passed to common Deep Learning based classification architectures for the second step.

One-Step Object Detection algorithms combine the detection and classification step by introducing the idea of regressing the bounding box predictions. These algorithms are the go-to choices for most of the real-time object detection tasks due to their speed of detection.
  
Neural networks have been used for the identification of unlabelled data based on labelled training data~\cite{svozil1997introduction}. 
Further improvements have been seen with the introduction of convolutional neural networks~\cite{he2016deep} in terms of overall prediction. They have been used for plant classification~\cite{Hassan:2021,toth2016deep}.  Previous work has examined using machine learning with deep neural networks for general fruit classification using EfficientNet~\cite{DUONG2020105326}. Results indicate deep neural networks offer superior prediction abilities but the increased performance comes with higher computation cost.  Detection, classification and mapping systems have been developed for coffee assessment on branch~\cite{RAMOS20179} and during harvest~\cite{BAZAME2021106066}.  Previous work has examined the assessment of fruit on trees for improvements on harvesting~\cite{CherryOnTree:2020}.  To our knowledge, there is no published work using deep learning to improve the assessment and classification of cherry phenotyping in cherry breeding programs.  
  
Different object detection models proposed have their own strengths and weaknesses. Commonly investigated models are the HAAR Cascade Classifier~\cite{HAAR:2001}, Region-Based Convolutional Neural Networks (R-CNN)~\cite{R-CNN:2017}, the Single Shot Detector (SSD)~\cite{ssd:2016} and You Only Look Once (YOLO)~\cite{yolo:2016} models.  

\subsection{HAAR Cascade Classifier}

HAAR Cascade is a machine learning object detection algorithm used to identify objects in an image based on the concept of features~\cite{HAAR:2001}. A cascade function is trained from a large collection of positive and negative images.   Once trained, it is then used to detect objects in other images. The algorithm has four stages:
\begin{itemize}
    \item HAAR feature selection
    \item Creating integral images
    \item Adaboost training
    \item Cascading classifiers
\end{itemize}

This method is well known for detecting faces and body parts in an image and can be trained to identify almost any object. The algorithm needs a large number of positive images of cherries and negative images without cherries to train the classifier. The model is then trained while implementing the four steps.

\subsection{Faster R-CNN}
R-CNN models~\cite{girshick2014rich} are two-step object detection models known as \emph{Regions with CNN Features} or \emph{Region-Based Convolutional Neural Network}.  The R-CNN model is comprised of three steps incorporating:

\begin{itemize}
    \item Region Proposal: Generate and extract category independent region proposals, e.g. possible bounding boxes
    \item Feature Extractor: Extract features from all the proposed regions in the first step using a deep convolutional neural network
    \item Classifier: Classify features as one of the known classes
\end{itemize}

The R-CNN and Fast R-CNN~\cite{girshick2015fast} use selective search to find the region proposals. Selective search is a slow and time-consuming process affecting the performance of the network. Faster R-CNN~\cite{R-CNN:2017} was introduced that eliminates the selective search algorithm and lets the network learn the region proposals. For the Faster R-CNN model the image is provided as an input to a convolutional network that provides a convolutional feature map. Instead of using a selective search algorithm
on the feature map to identify the region proposals, a separate network is used to predict the region proposals. The predicted region proposals are then reshaped using a RoI (Region of Interest) pooling layer that is then used to classify the image within the proposed region and predict the offset values for the bounding boxes.

The architecture of the model takes a set of region proposals as input from the image that are passed through a deep convolutional neural network. The end of the deep CNN is a custom layer called RoI Pooling, which extracts features specific for a given input proposed regions. The model splits into two outputs, one for the class prediction step involving a softmax/sigmoid layer, and another with a linear output for the bounding box. This process is then repeated multiple times for each region of interest. Faster R-CNN is considerably faster than R-CNN and Fast R-CNN and is a potential option for real time object detection.

\subsection{Single Shot Detectors}

Unlike the previous models of object detection which involved having one part of the network dedicated to providing region proposals followed by a high-quality classifier to classify these proposals in the second part, Single Shot Detector (SSD) methods are very accurate but come with a drawback of large computational cost (low frame rate). They are not suitable for real-time applications.

An alternate strategy for object detection can be accomplished  by combining these two tasks into one network.  Rather than having a network provide region proposals, a model can have a set of predefined boxes to identify objects. Using convolutional features maps from later layers of a network, the model can use small CONV filters over these features maps to predict class scores and bounding box offsets. As the model only requires one single shot to detect multiple objects within the image, this model was termed as Single Shot Detectors.
In the final layers, each pixel represents a larger area of the input image and this is used to infer the object position. 
SSD has two components: a backbone model and SSD Head.  The SSD head is just one or more convolutional layers added to the backbone and the outputs are interpreted as the bounding boxes and classes of objects in the spatial location of the final layer activations.

\subsection{You Only Look Once (YOLO)}

YOLO~\cite{yolo:2016} is a convolutional neural network (CNN) for performing object detection tasks in real-time. The algorithm applies a single neural network to the full image, and then divides the image into regions and predicts bounding boxes and probabilities for each region. These bounding boxes are weighted by the predicted probabilities. The single neural network simultaneously predicts multiple bounding boxes and class probabilities for those boxes. YOLO trains on full images and directly optimizes detection performance. This model is extremely fast and learns generalizable representations of objects and outperforms its competitor models.
YOLOv3~\cite{redmon2018yolov3} uses a variant of Darknet, which originally has a 53-layer network trained on ImageNet that offers significant improvements over previous YOLO models. For detection, 53 more layers are stacked onto it, forming a 106 layer fully convolutional architecture.  

\section{Model Development and Data}\label{sec:modelDev}

\begin{figure}[!tb]
\centerline{\includegraphics[width=0.8\linewidth]{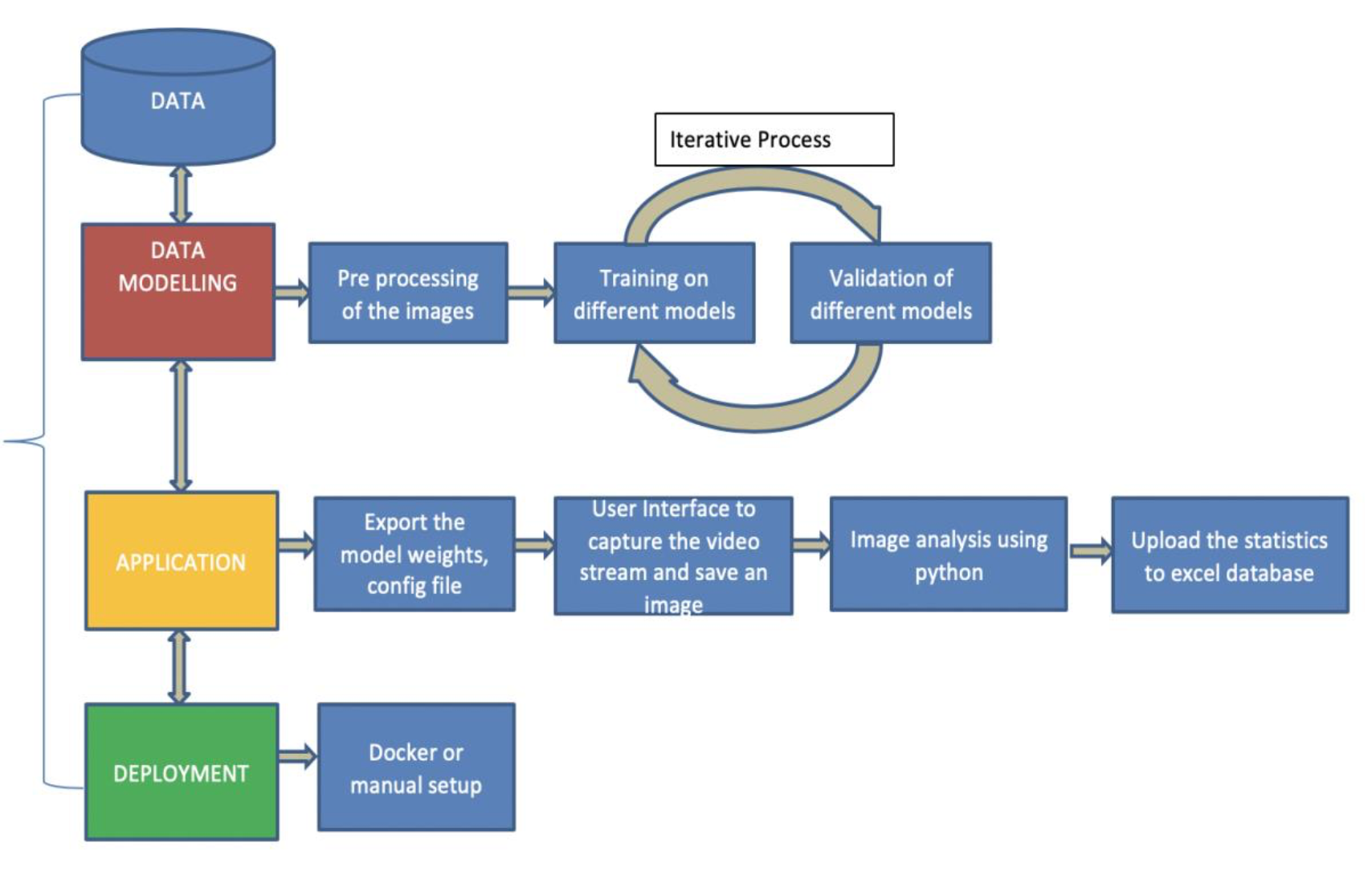}}
\caption{Workflow diagram}
\label{fig:workflow}
\end{figure}

A variety of tools, software and frameworks were used for building the application and interface for the detection of cherries.  Figure~\ref{fig:workflow} describes the complete workflow. The project was developed with Python. The system contains:

\begin{itemize}
    \item OpenCV\footnote{\url{https://opencv.org/}}: an open source project written in C++ with bindings for various other languages (Python, Java) (OpenCV version 4.3).
    \item TensorFlow\footnote{\url{https://www.tensorflow.org}}: a deep learning library developed and maintained by Google. TensorFlow has primarily been used to train the model, define the architectures of different deep learning models and validate the performance of the model. The SSD and Faster R-CNN models were trained using TensorFlow. 
    \item Tkinter\footnote{\url{https://wiki.python.org/moin/TkInter}}: a Python library that helps with Graphical User Interface development (GUI) for desktop applications.
    \item Google Colab\footnote{\url{https://colab.research.google.com}}: a free virtual environment provided by Google with different virtual machines powered by GPUs that enable training and development of deep learning models. All models were trained with Google Colab.
    \item LabelImg\footnote{\url{https://github.com/tzutalin/labelImg}}: a Python-based library to manually annotate the images.
\end{itemize}

Agriculture and Agri-Food Canada provided test images. 360 raw JPEG images containing several cherries in addition to the label cards to help in classifying the cherries to a category were analyzed. A ruler (scale) was also placed on each image that gives the approximate size of each cherry.

Pre-processing the images involved manually annotating the images using the LabelImg library. A bounding box was drawn around each cherry in the image. After drawing the boxes, the image is saved either in the YOLO format (an XML format) or PASCAL VOC format (a text file). After manual annotation of images, the data is usable for all of the deep learning architectures.

\subsection{Model Training}

Training different deep learning architectures on the data is an iterative process. The process involves configuring the deep learning architecture, continuously tuning the hyperparameters, measuring loss after each epoch to see if it is improving and finally monitoring the metrics such as the accuracy, the number of false positives and false negatives. For this application, the data was used to train three different deep learning architectures using YOLOv3, Faster R-CNN and SSD. For SSD and Faster R-CNN, the models were trained using TensorFlow.  Both the  YOLOv3 base model and YOLOv3 with Efficient Net~\cite{Tan:2019} as the backbone model were trained for comparison.   

\subsection{Model Performance Evaluation Metrics}

When comparing the performance of the models, the Intersection over Union (IoU)~\cite{IoU} and average precision and accuracy were considered. Bounding box predictions are not very precise on the pixel level, and thus a metric is required for the extent of overlap between two bounding boxes (true and predicted boxes). Intersection over Union takes the area of intersection of the two bounding boxes and divides it with the area of their union (Equation~\ref{eqn:iou}) as:

\begin{eqnarray}\label{eqn:iou}
    \text{IoU}&=& \frac{\text{Area of Overlap}}{\text{Area of Union}.}
\end{eqnarray}

IoU produces a score between 0 and 1 that represents the quality of overlap between the two boxes. A score of 1 means a perfect match between the two boxes and 0 means no match between the two boxes. 

Precision provides information on how accurate the model predictions are, and recall denotes whether it can detect all objects present in the image. Average Precision (AP) is a widely used metric in object detection evaluation. Accuracy in counting cherries in the image was used to measure model performance.

\subsection{Model Selection}

\begin{figure*}[!t]
\captionsetup[subfigure]{justification=centering}
\begin{subfigure}[t]{0.5\textwidth}
\centering
\includegraphics[width=.9\textwidth]{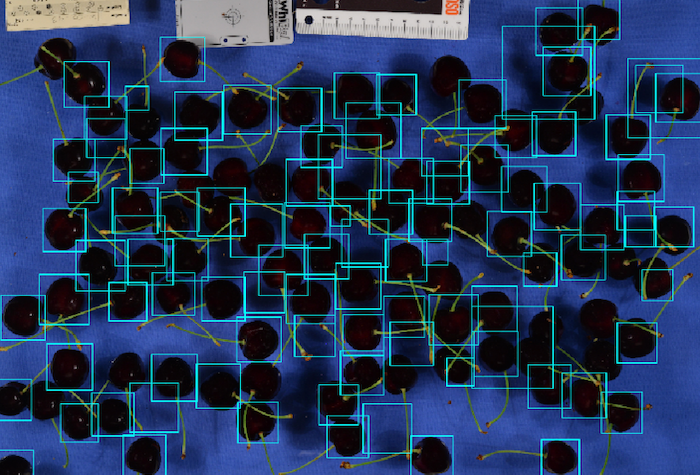}
\caption{HAAR model detection. }
\label{fig:haar}
\end{subfigure}
\hfill
\begin{subfigure}[t]{0.5\textwidth}
\centering
\includegraphics[width=.9\textwidth]{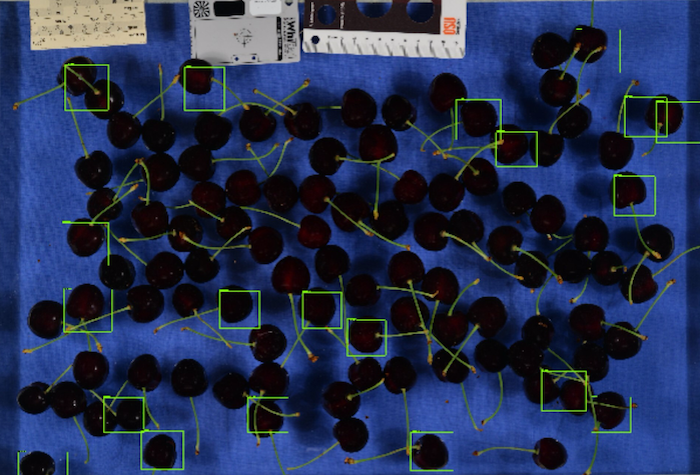}
\caption{SSD model detection}
\label{fig:ssd}
\hfill
\end{subfigure}
\begin{subfigure}[t]{0.5\textwidth}
\centering
\includegraphics[width=.9\textwidth]{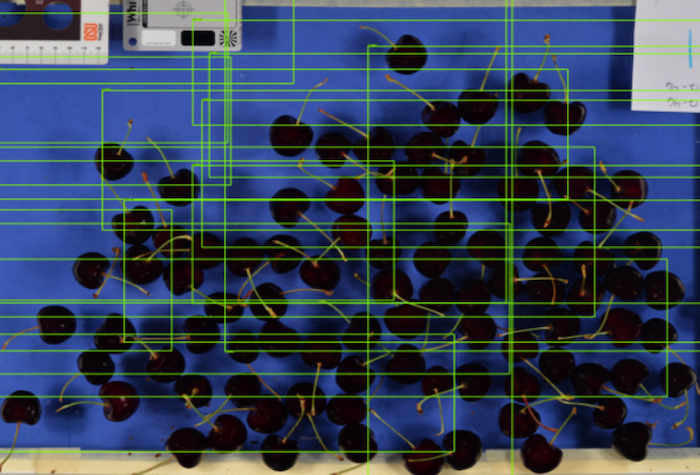}
\caption{R-CNN model detection}
\label{fig:r-cnny}
\end{subfigure}
\hfill
\begin{subfigure}[t]{0.5\textwidth}
\centering
\includegraphics[width=.9\textwidth]{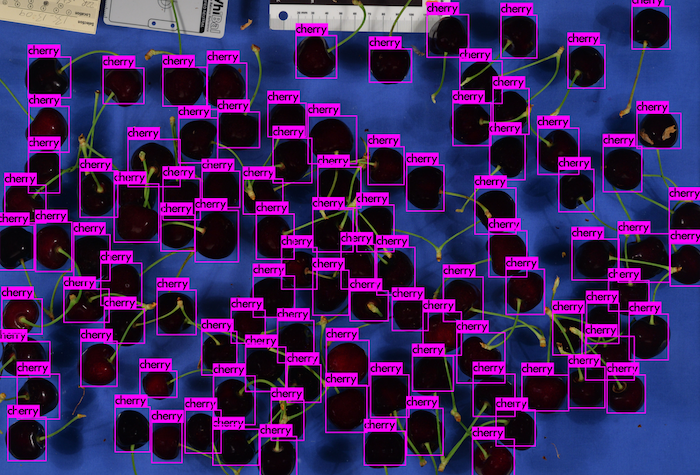}
\caption{YOLO model detection}
\label{fig:YOLO}
\end{subfigure}
\caption{Prediction results of the four models on the sample cherry image from test set using different modelling techniques}\label{fig:viz-model}
\end{figure*}

Four models were trained with different techniques to identify the most suitable modeling architecture for our custom cherry dataset: the HAAR – Feature Cascade Classifier, Faster R-CNN, Single Shot Detectors (SSD) and You Only Look Once (YOLO) models.  The object detection results were analyzed on a testing set to identify which technique would be most suitable for the cherry image set.

Figure~\ref{fig:haar} presents a sample image for the HAAR model.  While the model is computationally efficient on prediction, the training is slow and accuracy is low. The model detects shadows as cherries as well as demonstrating a high variance based on image size and offers no multi-class detection.

Figure~\ref{fig:ssd} presents a sample for the SSD model. This neural network architecture takes a single shot to detect multiple objects within the image and combines region identification and class prediction steps into a single neural network. However, it fails to detect most of the objects.  Additionally, the bounding boxes are not precise in terms of cherry size and location.  

Figure~\ref{fig:r-cnny} presents a sample for the R-CNN model using a 2-step approach to detect objects. For the cherry image set, the detection accuracy is very low.  Further, it requires a large training data set and is computationally expensive.  

Figure~\ref{fig:YOLO} presents a sample image for the YOLO model.  While it utilizes a similar underlying concept to SSD models,  YOLO uses a different neural network architecture.  It is computationally efficient in real-time prediction. The model has a very high detection accuracy and precise detection of bounding boxes.  

From the comparison of the different models utilizing the cherry image set, the results indicate that the YOLO model is the suitable choice for cherry object detection.  This model not only correctly identifies almost all of the cherries in the image but also provides fairly accurate bounding boxes for each cherry detected. Detailed results from subsequent training of YOLO models are in Section~\ref{sec:Analysis}.

\subsection{Application Development}

The end-user application has a user interface and interactivity with Excel. The trained model was incorporated into the application with OpenCV and implements two steps:

\subsubsection{Video Stream and Image Capture}

\begin{figure*}[!t]
\centerline{\includegraphics[width=1\textwidth]{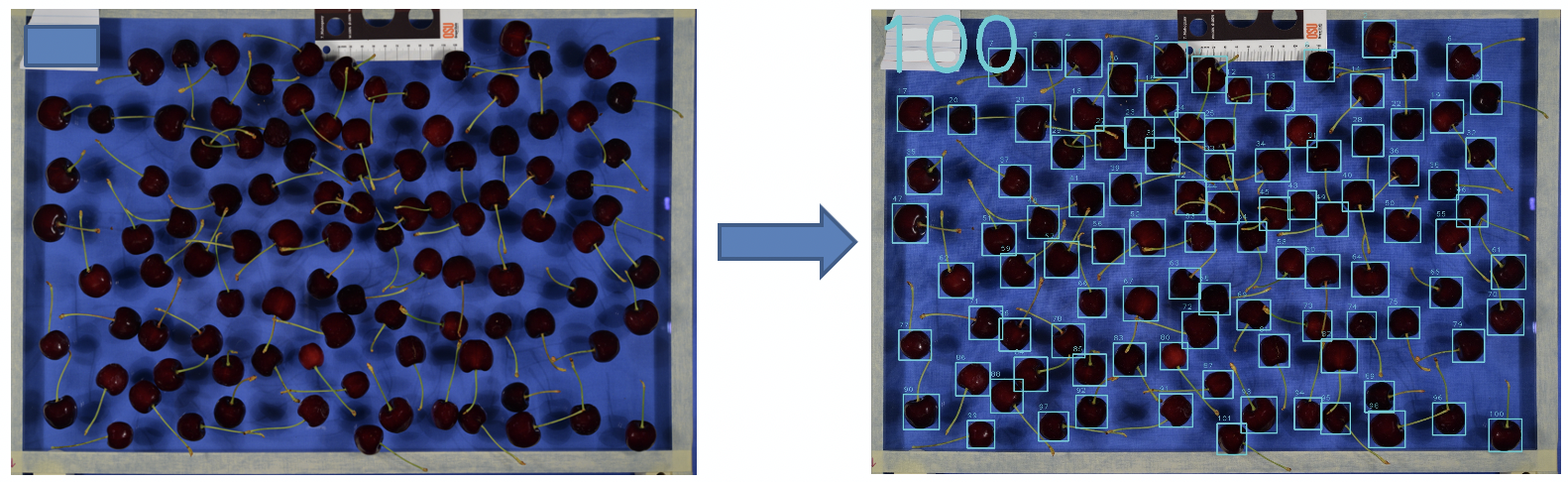}}
\caption{Application showing input and output images}
\label{fig:application}
\end{figure*}

The application consists of switching on the USB web camera from OpenCV. Once the camera is switched on, each frame from the video is taken and passed to the YOLOv3 original and Efficient Net models.  As each frame from the video is passed to the model, the model analyzes the frame and displays the corresponding bounding boxes on the screen where cherries are detected and the overall cherry count. Figure~\ref{fig:application} shows the display output. The reading at the top left is the cherry count from the YOLOv3 model.

Once the operator is satisfied with the count converging for both models, pressing the space bar generates a pop-up window for the operator allowing them to save the image from the video for additional analysis. The operator can enter the specific name of the genotype for that batch of cherries, and the image is saved in the images folder for that genotype and annotated with a timestamp.

\subsubsection{Extracting Feature Data}

The second part of the application consists of two scripts. The first script stores each image in the images folder and generates the predicted image with the bounding box using the YOLOv3 model. A CSV file is also created that gives confidence and bounding box coordinates for each cherry. This information is essential to crop each cherry from the actual image to get its size, color and various other properties. The second script is used to take each image in the images folder and run through a series of functions to populate the results spreadsheet. The size is measured using the largest dimension if it is a rectangle. The color classification is based on extracting the average red, green and blue content of each image and then comparing them with how close they are to the average intensity of different classes.  This results in the generation of four spreadsheets.  Table~\ref{tab:application} summarizes the information stored in each spreadsheet.  

\begin{table}[!t]
\caption{Information captured from the application}
\label{tab:application}
\centering
\begin{tabular}{c|c}
\hline
\textbf{Sheet} & \textbf{Details}                                                  \\ \hline
Summary        & \begin{tabular}[c]{@{}c@{}}ImageID, Count, Avg size, Avg size - Top 50, \\ Average red/green/blue content, Average RGB content - Top50, \\ Stem Avg Red, Stem Avg Green, Stem Avg Blue, Date and Time\end{tabular} \\ \hline
Cherry Size    & \begin{tabular}[c]{@{}c@{}}ImageID, CherryID, Confidence, Cherry Size, Cherry Width/Height, \\Cherry Size (mm), Cherry Width/Height (mm), Top 50, \\ Box X/Y, Central Region, Scaled Box X/Y, Date, Time\end{tabular}                          \\ \hline
Cherry Colour  & \begin{tabular}[c]{@{}c@{}}Image ID, Cherry ID, Cherry Avg Red, Cherry Avg Green,\\Cherry Avg Blue,  Cherry Classification, Date, Time, Top 50\end{tabular}                                                                                                                               \\ \hline
Stem Colour    & \begin{tabular}[c]{@{}c@{}}Image ID, Cherry ID, Stem Avg Red, Stem Avg Green, \\ Stem Avg Blue, Date, Time, Top 50\end{tabular}                                                                                                                                                           \\ \hline
\end{tabular}
\end{table}

The summary sheet provides a summary of the entire image with the count and other traits. The second sheet is specific to each cherry in an image. The third sheet corresponds to the color details of each cherry in an image and the last sheet gives details about each stem of a cherry in an image. This is accomplished by cropping each cherry from the image using the coordinates produced by the text files script and extracting its average red, green and blue intensity. The average size, color and class of each cherry and its stem is recorded.

The entire code is packaged into two executable batch files. The first batch file will launch OpenCV and run the camera and save the images as required by the  operator. The second batch file consists of the two scripts that are used to generate the text file for the images and update the Excel spreadsheet.

\section{Analysis and Results}\label{sec:Analysis}

\begin{table*}[htbt]
\caption{Counting and IoU accuracy for the YOLOv3 original using Darknet-53 and the YOLOv3 Efficient Net B0 model for the training image set}
\label{tab:trainingData}
\begin{center}
 \begin{threeparttable}
\begin{tabular}{c|c|c|c|c|c|c|c|c|c}
\hline
\textbf{Model}    & \textbf{Resize} & \textbf{CT} & \textbf{DC} & \textbf{TC} & \textbf{TP} & \textbf{FP} & \textbf{FN} & \textbf{mAP @ 0.5} & \textbf{mean IoU} \\ \hline
original          & 512x416         & 0.5         & 6539        & 6498        & 6498        & 1           & 0           & 99.98\%            & 94.98\%           \\ \hline
efficientnet\_B0 & 480x480         & 0.25        & 6496        & 6498        & 6495        & 0           & 3           & 99.94\%            & 94.21\%           \\ \hline
\end{tabular}

\begin{tablenotes}[para,flushleft]
{\small Notes: Confidence Threshold (CT), Detection Counts with confidence threshold less than 0.1 (DC), True Counts (TC), True Positive (TP), False Positive (FP), False Negative (FN), mean accuracy precision with confidence threshold 0.5 mean (mAP@0.5), and mean intersection over union (mean IoU).
}
\end{tablenotes}
\end{threeparttable}
\end{center}

\caption{Counting and IoU accuracy for the YOLOv3 original using Darknet-53 and the YOLOv3 Efficient Net B0 model with the validation image set}
\label{tab:validate}
\begin{center}

\begin{tabular}{c|c|c|c|c|c|c|c|c|c}
\hline
\textbf{Model}    & \textbf{Resize} & \textbf{CT} & \textbf{DC} & \textbf{TC} & \textbf{TP} & \textbf{FP} & \textbf{FN} & \textbf{mAP @ 0.5} & \textbf{mean IoU} \\ \hline
original          & 512x416         & 0.5         & 3482        & 3455        & 3455        & 0           & 0           & 99.97\%            & 95.16\%           \\ \hline
efficientnet\_B0 & 480x480         & 0.25        & 3452        & 3455        & 3451        & 0           & 4           & 99.88\%            & 94.26\%           \\ \hline
\end{tabular}

\end{center}

\end{table*}

Three questions were considered when testing how the models performed:

\begin{itemize}
    \item What is the counting and IoU accuracy of the models on static images?
    \item What is the counting accuracy and speed of the models in real time counting?
    \item Is the extracted color and size accurate compared with traditional methods?
\end{itemize}

Two YOLO-based models evaluated: YOLOv3 Efficient Net B0 model~\cite{Tan:2019} and YOLOv3~\cite{redmon2018yolov3} original. In total, 364 images are in the data set. For most of these images, the cherry count is targeted to be 100, but as these images involved workers manually counting 100 cherries some variability in total count is observed.  A total of 109 images out of these 364 images where randomly chosen and labelled using LabelImg. The models were trained with 72 randomly chosen images with the remaining 37 images used as a validation set. Finally, all models were tested with the complete 364 image set for performance comparison.

\subsection{Counting and IoU Accuracy}

Table~\ref{tab:trainingData} presents the results for YOLOv3 original and YOLOv3 Efficient Net B0 models for the 72 training images.  Table~\ref{tab:validate} presents the results for the models with the validation image set of 37 images.  Table~\ref{tab:allImages} presents the overall performance of the trained models on the complete image set.  

\begin{table*}[!t]
\caption{Counting accuracy for the YOLOv3 original using Darknet-53 and the YOLOv3 Efficient Net B0 model for complete image set}
\label{tab:allImages}
\begin{center}
 \begin{threeparttable}

\begin{tabular}{c|c|c|c|c|c|c|c}
\hline
\textbf{Model}    & \textbf{Resize} & \textbf{CT} & \textbf{TC} & \textbf{TP} & \textbf{FP} & \textbf{FN} & \multicolumn{1}{l}{Human Count Error} \\ \hline
original          & 512x416         & 0.5         & 34963       & 34961       & 1           & 2           & \multirow{2}{*}{22}                    \\ \cline{1-7}
efficientnet\_B0 & 480x480         & 0.25        & 34963       & 34953       & 2           & 10          &                                        \\ \hline
\end{tabular}

\begin{tablenotes}[para,flushleft]
{\small Notes: Confidence Threshold (CT), True Counts (TC), True Positive (TP), False Positive (FP), and False Negative (FN).
}
\end{tablenotes}
\end{threeparttable}
\end{center}
\end{table*}

For the testing results, consideration was also given for reasons why the models predicted the counts incorrectly. Of all 364 images, the YOLOv3 Efficient Net B0 model predicts counts incorrectly for 11 images and 12 cherries missed. The reasons for missing cherries is that some cherries are on the frame edges of the images, some leaves are counted as cherries, and a few completely failed to be detected.  The YOLOv3 model predicts three image counts wrong while incorrectly counting three cherries (one leaf counted as cherry and two cherries due to overlapping).  There are a total of 22 images that were manually counted wrong with not exactly 100 cherries.

From the training and validation results, it can be seen that both models achieved accuracy (based on AP metric) of over 99\%, and the predicted bounding box around cherry accuracy (based on mean IoU metric) are  over 94\%.  The results from the YOLOv3 original model perform slightly better in both metrics.  It was decided to use YOLOv3 original as the model to generate results when dealing with static images.  The testing results indicate that the models outperformed the traditional method of manually counting. The Efficient Net model predicted 11 wrong counts and the original model predicted only three wrong counts, while the manually counting method produced 22 wrong counts.  From the testing results, wrong counts by the models are due to either cherries are on the camera frame edge or the cherries are overlapping. Both of these scenarios can be avoided when displaying cherries real time in the future for the operator to view. Both of the models can be used on the real time counting application to check with each other to ensure that the count is correct.

\subsection{Real Time Counting Accuracy and Speed}

After training, validating, tuning and testing the models, an application was developed to use the models to perform the real time counting.  Models were tested with several videos provided by AAFC mimicking the lab real time counting situation. It takes approximately 60 seconds to manually count 100 cherries. With the application, utilizing both models at the same time, the time to process the image and perform the count is approximately one second utilizing a consumer laptop. The YOLOv3 original model takes around 0.5 seconds to finish one inference, and the YOLOv3 Efficient Net B0 model takes around 0.1 seconds. 

\subsection{Color and Size Information Extraction}

\begin{table}[!t]
\begin{minipage}[c]{\textwidth}
\centerline{\includegraphics[scale=0.5]{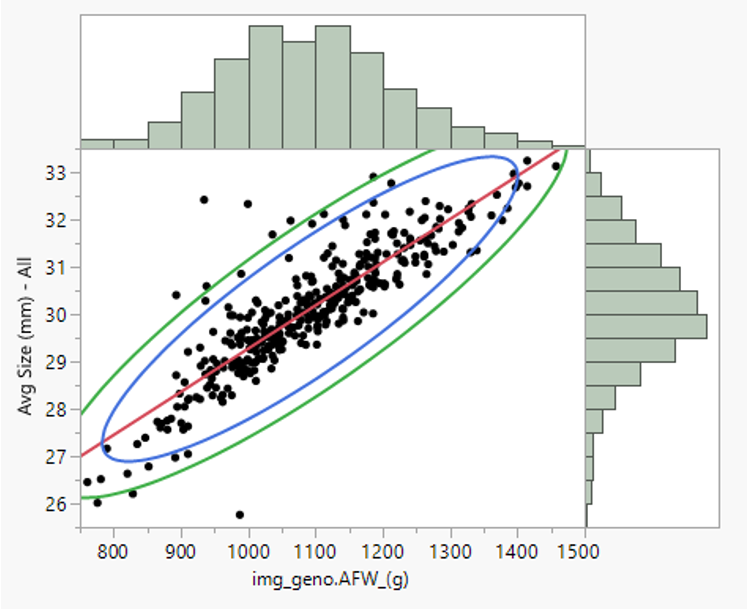}}
\caption{Bivariate fit of YOLOv3}
\label{fig:bivariate}
\end{minipage}
\hfill
\begin{minipage}[c]{\textwidth}
\centering
\caption{Summary Statistics for Bivariate fit of YOLOv3}
\label{tab:SummaryStat1}
\begin{subtable}{0.75\textwidth}
\resizebox{\columnwidth}{!}{%
\centering
\begin{tabular}{c|c|c|c|c}
\hline
\textbf{}   & \textbf{Value}                & \textbf{Lower 95\%}   & \textbf{Upper 95\%}   & \textbf{Signif. Prob} \\ \hline
Correlation & 0.874896                      & 0.848403              & 0.897017              & \textless{}.0001*     \\ \hline
Covariance  & \multicolumn{1}{l|}{145.1858} & \multicolumn{1}{l|}{} & \multicolumn{1}{l|}{} & \multicolumn{1}{l}{} \\ \hline
Count       & 364                           &                       &                       &                       \\ \hline
\end{tabular}
}
\end{subtable}
\hfill 
\begin{center}
\begin{subtable}{\textwidth}
\centering
\begin{tabular}{c|c|c}
\hline
\textbf{Variable}  & \textbf{Mean}                 & \textbf{Std Dev}              \\ \hline
img\_geno.AFW\_(g) & 1092.03                       & 125.8303                      \\ \hline
Avg Size (mm) - Al & \multicolumn{1}{l}{30.10884} & \multicolumn{1}{l}{1.315497} \\ \hline
\end{tabular}
\end{subtable}
\end{center}
\end{minipage}
\end{table}

After verifying cherry counts using the models, a static image is captured. This image is then passed to YOLOv3 original model to get another inference to get the bounding box information for each cherry in the image.  Based on the bounding box coordinates, the model generates information on the color, cherry size, and stem color. This information is stored in an Excel spreadsheet for additional analysis.  

To ensure that the two models produce color and size information suitable for use in plant phenotyping, the 364 static images were analyzed using this method.  The results were compared to the results generated by traditional methods.

To compare how much the proposed methods and the traditional method agree with each other, a bivariate plot, correlation analysis, and linear regression were conducted.  The results from YOLOv3 original and YOLOv3 Efficient Net are very similar, so only the results from YOLOv3 original are shown in Figure~\ref{fig:bivariate}.  Table~\ref{tab:SummaryStat1} contains summary statistics comparing YOLOv3 original with the traditional method. In the plot \emph{Avg Size (mm)-All} is the predicted results from YOLOv3 original and \emph{img\_geno.AFW\_(g)} is from the traditional method.

In the bivariate plot, the two variables correlated quite well with very few outliers. This indicates that the proposed method and the traditional method agree with each other in the estimation of cherry size.  The correlation value is 0.875, and p value is  0.001, which indicates that these
two variables are highly positive correlated. The linear fit shows the relationship between the predicted size and the weight from the traditional method. This linear relation explains 76.5\% of the total variance in the data.

The reason why the two variables are not completely agreeing with each other was examined.  The two variables calculated are not based on the same sample from the same population (tree). The proposed method uses 100 cherries in the image while the weight variables are calculated from two batches of around 100 cherries which includes the 100 cherries in the image.  The larger value of the bounding box width and height is used to estimate the size of the cherry in the proposed method, while the traditional method uses average weight to estimate the size.  Additionally, the fisheye effect is not taken into consideration.   As discussed previously, there are also manual count or prediction errors.

\begin{table}[!t]
\caption{Average colour prediction for YOLO-based models}
\label{tab:colour}
\begin{center}
\begin{tabular}{c|c|c|c}
\hline
\textbf{Models}        & \textbf{Total Counts} & \textbf{Correct Counts} & \multicolumn{1}{l|}{\textbf{Accuracy}} \\ \hline
YOLOv3\_original       & 364                   & 262                     & 72\%                                   \\ \hline
YOLOv3\_efficient\_net & 364                   & 261                     & 72\%                                   \\ \hline
\end{tabular}
\end{center}
\end{table}

Table~\ref{tab:colour} presents the average colour prediction for each of the YOLO models.  The analysis was based on whether the average predicted colour type falls in the range of the ones provided by the traditional method.

The results shows that the accuracy is  72\%. Considering the sample difference and the variance in the human judgement of the colour, these results are acceptable based on input from subject matter experts. The possible reasons causing this discrepancy includes the colour scores given by technicians are subjective and are impacted by lighting conditions as well as an individual's unique visual colour perception characteristics.  Further, sample inconsistency, and image and colour card image white balance issues can further impact assessment. 

\section{Conclusion and Future Work}

New cultivars generate enormous economic value for the horticulture industry. As a critical step in the creation of new cultivars, pheotyping requires a significant investment in terms of time and resources, especially when humans perform tedious and repetitive tasks such as data collection. A Python-driven application utilizing YOLO was developed to extract critical information from images including the number of cherries, the size and the colour of each cherry. This application helps to modernize the existing phenotyping procedures, permitting evaluation on a much larger scale and generation of accurate and reproducible data during research.  

The YOLO models obtained close to 99\% accuracy in object detection and counting of cherries. They also scored over 90\% on the IOU metric indicating the model is both accurate in detecting cherries and very precise for object localization, which is important when extracting size and colour information from the detected objects. The model surpassed human performance and even detected manual errors on previous years image data set, which was used for training and testing the model. The model offers significant performance improvements in time, and processing with the model takes 95\% less time compared to manually counting. This research applies to many other fruit applications beyond cherries, and can be used by growers and environmental agencies.

Future work will examine further tuning of
parameters such as resizing images (regardless of whether using randomly resizing in the training process or not), confidence threshold, and non-max suppression threshold after training in order to improve the overall results.

\section{Acknowledgment}

The authors gratefully acknowledge the assistance of summer term students (Jakob Lavioe, Trista Algar) and technical staff (Chris Pagliocchini, Michael Wiess and Melanda Danenhower) of the Summerland Research and Development Centre of Agriculture and Agri-Food Canada (AAFC) in capturing and labelling the images used for training of models.

\bibliographystyle{spmpsci}
\bibliography{ref}

\end{document}